\documentclass[a4paper, 10pt, conference]{ieeeconf}      

\IEEEoverridecommandlockouts                              

\usepackage{graphicx} 
\usepackage{multirow}
\usepackage{cite}
\usepackage{subfigure}

\title{\LARGE \bf
Panoramic Annular Localizer: Tackling the Variation Challenges of Outdoor Localization Using Panoramic Annular Images and Active Deep Descriptors
}

\author{Ruiqi Cheng$^{1}$, Kaiwei Wang$^{1}$,  Shufei Lin$^{1}$, Weijian Hu$^{1}$, Kailun Yang$^{1}$, \\Xiao Huang$^{1}$, Huabing Li$^{1}$, Dongming Sun$^{1}$ and Jian Bai$^{1}$ 
\thanks{$^{1}$All of the authors are with the State Key Laboratory of Modern Optical Instrumentation, Zhejiang University, Hangzhou, China.
        {Corresponding author: Kaiwei Wang, \tt\small wangkaiwei@zju.edu.cn}}%
}

\begin{document}

\maketitle
\thispagestyle{empty}
\pagestyle{empty}

\begin{abstract}

Visual localization is an attractive problem that estimates the camera localization from database images based on the query image. It is a crucial task for various applications, such as autonomous vehicles, assistive navigation and augmented reality. The challenging issues of the task lie in various appearance variations between query and database images, including illumination variations, dynamic object variations and viewpoint variations. In order to tackle those challenges, Panoramic Annular Localizer into which panoramic annular lens and robust deep image descriptors are incorporated is proposed in this paper.  The panoramic annular images captured by the single camera are processed and fed into the NetVLAD network to form the active deep descriptor, and sequential matching is utilized to generate the localization result. The experiments carried on the public datasets and in the field illustrate the validation of the proposed system.

\end{abstract}

\section{INTRODUCTION}
Localization is one of important research topics concerning autonomous vehicles~\cite{ETHlandmark, ABLE}, robotics~\cite{Tang2019} and assistive navigation~\cite{KeyPosition, VisualLocalizer}. Generally, GNSS (global navigation satellite system) is the straightforward way to localize the autonomous vehicles in the urban areas. However, localization tends to fail at those fade zones, such as the streets with high-rises, or under severe conditions, such as bad space weather. Fortunately, as the non-trivial sensing source, visual information is eligible for providing localization tasks with sufficient place fingerprint. The proliferation of computer vision has spurred the researchers to propose a great deal of vision-based localization solutions~\cite{VPR_SURVEY, ETHlandmark, ABLE, Chemnitz, Bonn2019, FukuiITSC2018, MOLP}.
Given a query image, visual localization predicts the camera location by searching the best-matching database images featuring the largest similarity to that query image. 

The most challenging and attractive part of visual localization is that the appearance variations between query and database images impact on the similarity measurement of images, so as to impede the robustness of the algorithm. Those appearance variations involve illumination variations, season variations, dynamic object variations and viewpoint variations. 
The variations of illumination, season and dynamic objects have been thoroughly researched by the research community. On the contrary, viewpoint variations are highly related to camera FOV (field of view) , and are tough to tackle merely using ordinary cameras.  Therefore, the expansion of FOV is essential to overcome viewpoint variations between query images and database images. 

\begin{figure}
    \centering
    \includegraphics[width=\columnwidth]{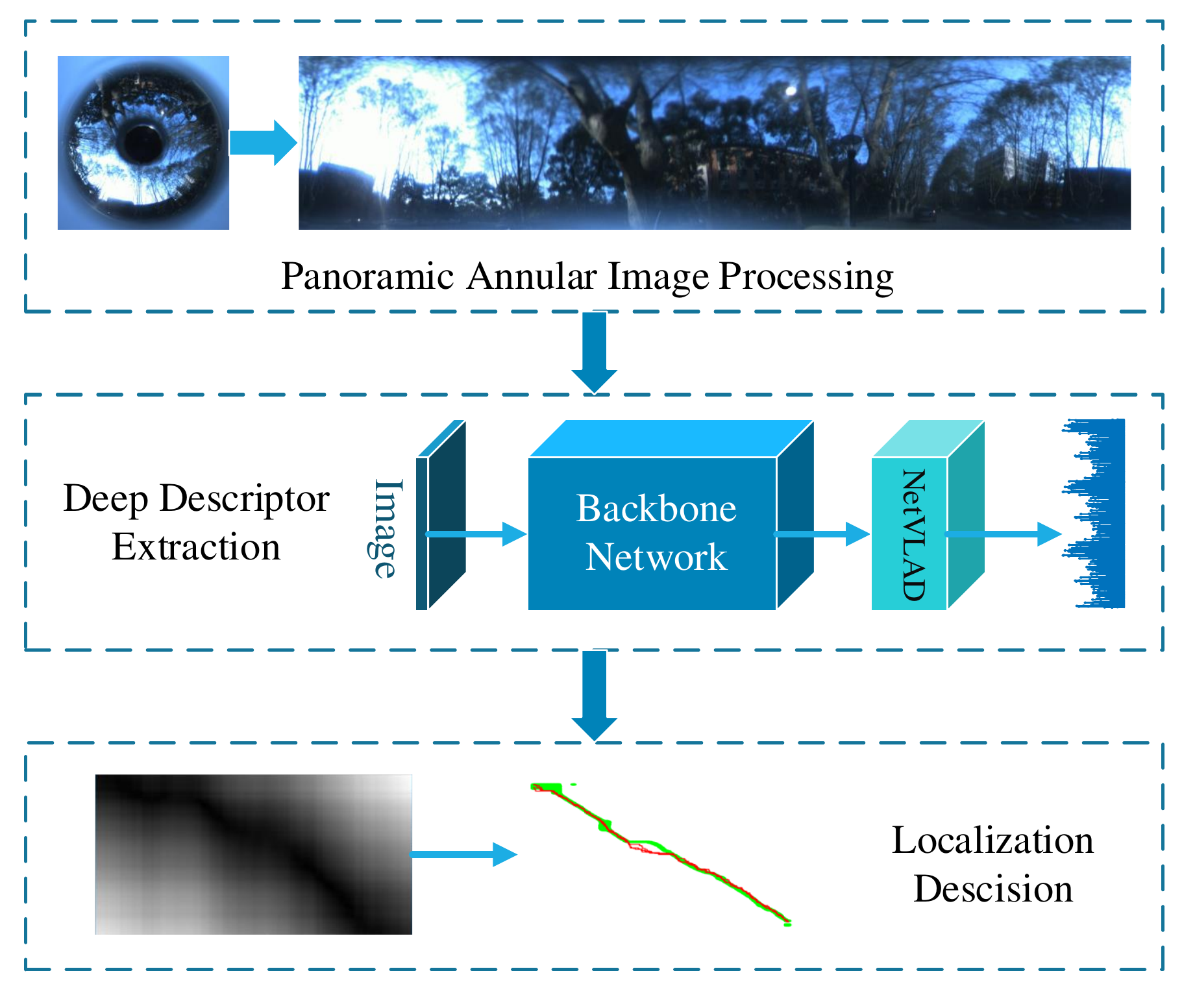}
    \caption{The schematic diagram of the proposed Panoramic Annular Localizer.}
    \label{fig:System}
\end{figure}

Based on our preliminary research~\cite{KeyPosition, VisualLocalizer, OpenMPR}, we propose a visual localization system PAL (panoramic annular localizer), which utilizes panoramic annular lens as the imaging sensor to capture the omnidirectional images of the surrounding environments and utilizes deep image descriptors to overcome various appearance changes. As shown in Fig.~\ref{fig:System}, the proposed PAL is composed of three phases: \textit{panoramic image processing}, \textit{deep descriptor extraction} and \textit{localization decision}.
The contributions of this paper are summarized as follows.
\begin{itemize}
\item In this paper, the panoramic annular lens is firstly integrated into outdoor visual localization system, which assists in the issues of viewpoint variations. Moreover, the car-mounted panoramic annular datasets captured in the real-world scenarios are released for the tasks of visual localization.
\item Active image descriptors extracted from panoramic images are leveraged to measure the similarity between images featuring various appearance changes. The proposed active descriptor outperforms the passive descriptors derived from feature maps of convolutional neural networks.
\item The active deep descriptors are combined with the sequential matching scheme. The performance of the proposed PAL system is validated on public and self-captured panoramic datasets for outdoor visual localization.
\end{itemize}


\section{Related work}
In this section, the prevailing panoramic cameras and visual localization solutions based on those cameras are summarized.
\subsection{Panoramic Cameras}
Various imaging systems are eligible to capture panoramic or omnidirectional images, including \textit{multi-camera systems}, \textit{catadioptric cameras} and \textit{panoramic annular cameras}.  
Those multi-camera systems, capture high-quality panoramic images, but the different exposure between different cameras may result in the illumination inconsistency in the panoramic image. A catadioptric camera usually constitute of an ordinary camera and a curved reflection mirror above the camera. 
Therefore, the mechanical structure is relatively complicated and the volume of camera is large compared with other types of panoramic cameras. 

Compared with those cameras, the panoramic annular lens~\cite{pal_Niu, pal_huang, pal_Zhou, pal_Luo, pal_huangX} features a simpler structure and lower cost. Through special optical design, the lateral light of lens is collected to the camera as much as possible by reflection twice within the panoramic annular lens. In view of the integrated package, the panoramic annular lens features flexible installation and is free from occlusion caused by camera's structure, especially compared with the catadioptric cameras.  Apart from that, the annular images get rid of the upper and lower part of spherical images and focus more on the intermediate part images that contain important visual cues of places.

\subsection{Image Features}
Extracting robust features from images is the fundamental factor that impacts the performance of visual localization. The research community has focused on this topic for a long time. 
As the early research on vehicle visual localization, SeqSLAM~\cite{OpenSeqSLAM2.0} utilized the SAD (sum of absolute difference) of normorlized image patches to measure the similarity between query and database images, which is not robust against various appearance variations. 
Aggregating local features (such as SURF~\cite{SURF} and ORB~\cite{ORB}) into compact vectors, BoW (bag of words) places a vital role in place recognition~\cite{OpenFABMAP, ORB-SLAM2}, especially becomes a popular place recognition approaches in SLAM (simultaneous localization and mapping) system. Unfortunately, in view of the limited description capability of hand-crafted local features, BoW performs badly under the conditions of illumination variations. 

Apart from the hand-crafted descriptors mentioned above, the feature maps of prevailing CNN (Convolutional Neural Network) are also leveraged as powerful images descriptors~\cite{Garden,VisualLocalizer}. 
The ``passive'' deep descriptors cope with variation  issues by the limited pretrained knowledge derived from vanilla classification tasks, which results in sub-optimal localization performance  when compared with the active deep descriptors. There are several work~\cite{netVLAD, activeCNN, netVLADpano} that falls into  active image descriptors, which are trained specially to robustify the descriptor performance under the conditions of appearance variations. 

\subsection{Panoramic Visual Localization}
In order to tackle the viewpoint variations, the research community has proposed various visual localization approaches to achieving robust place recognition. Murillo and Josecka proposed place recognition using GIST descriptors extracted from panoramic images~\cite{gist_panoramas}. As one of the earliest attempts on the task, the proposed approach achieved good performance on a large-scale dataset, but the issues of appearance variations are not attached importance to. Based on the NetVLAD module,  the image retrieval approach of panoramic images proposed in~\cite{netVLADpano} performed well on the street view dataset. However, the algorithm was not designed for the localization problems and not validated on other  panoramic datasets. Oishi et al~\cite{SeqSLAM++} proposed view-based robot localization and navigation, where panoramic images are one of the multi-modal data. The hand-crafted image features and the sliding window scheme were utilized for matching the panoramic images, which naturally behaves inferior when matching images with apparent variations.

\section{Methodology}
In this section, we elaborate the proposed PAL system, a visual localization framework that is designed for the challenging variation issues of visual localization. Panoramic annular lens captures the omnidirectional images with $360^\circ$ FOV in single frame, which is the effective camera to tackle the viewpoint variations in visual localization. Apart from that, efficient deep image features are utilized to extract the place fingerprint embedded in images, which boosts the robustness against appearance variations, such as illumination, season and dynamic object variations.

\subsection{Preprocessing of panoramic annular images}
\begin{figure}
	\centering
	\subfigure[]{	
		\includegraphics[width=\columnwidth]{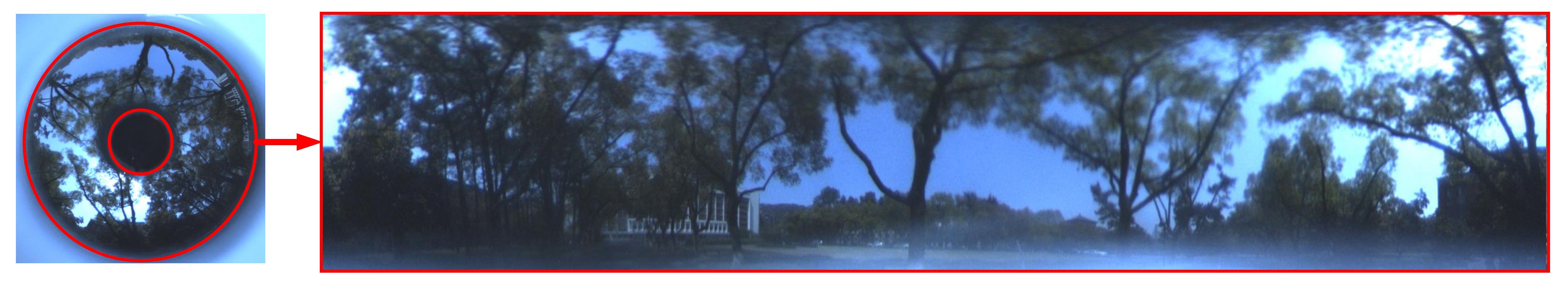}
	}
\subfigure[]{
	\includegraphics[width=\columnwidth]{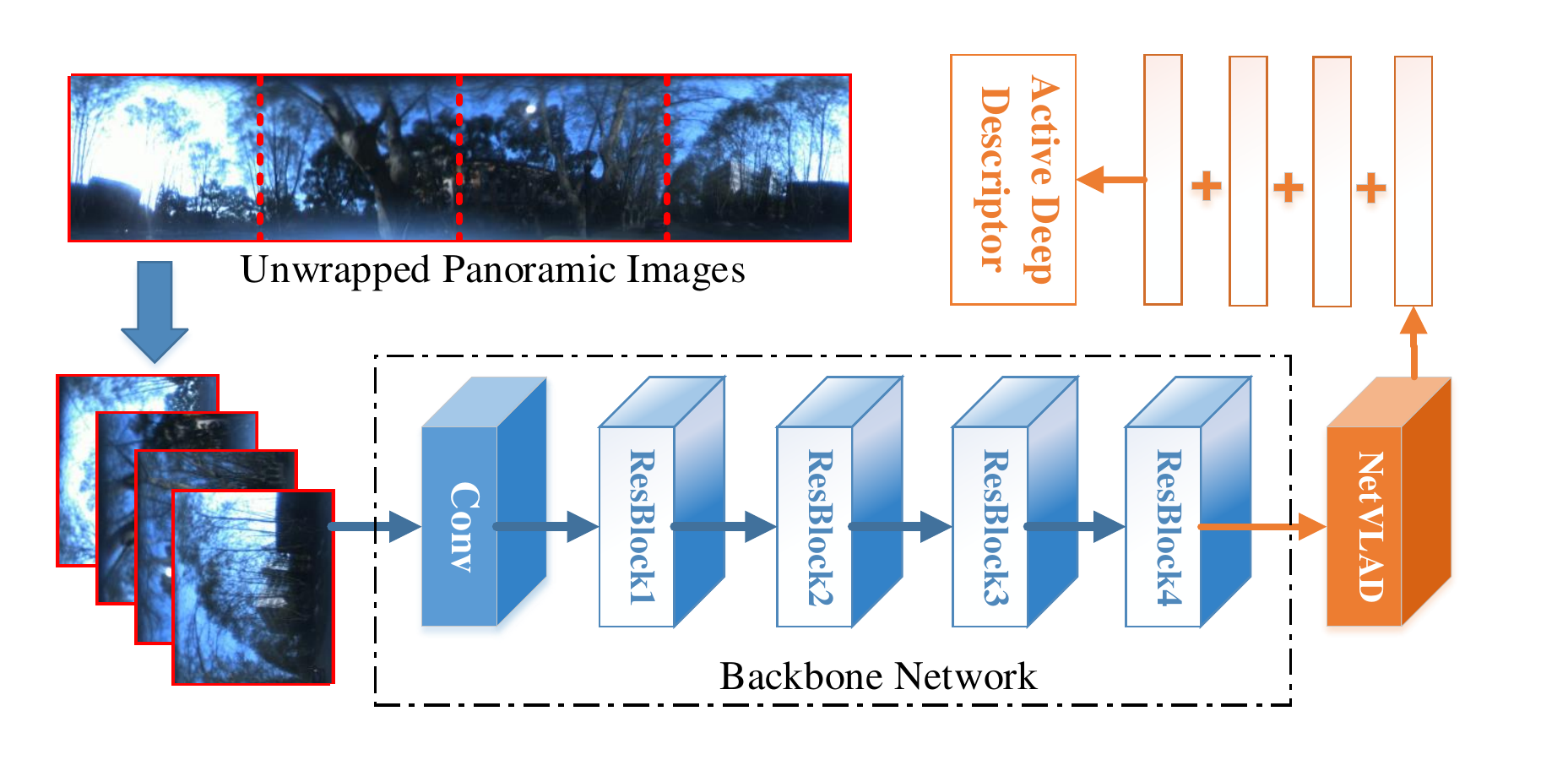}
}
	
\caption{The unwrapping process and feature extraction of panoramic annular images. (a) Unwrapping the panoramic annular image into the rectangular image. (b) The extraction procedures of the active deep descriptor from the panoramic image.}
\label{fig:ActiveDescriptor}	
\end{figure}

As the name suggests, panoramic annular images [e.g. the left image of Fig.~\ref{fig:ActiveDescriptor} (a)] are annular images that covers $360^\circ$ field of view. In order to apply different kinds of feature extraction procedures to the panoramic images, the annular images are firstly unwrapped into the rectangular images, shown as the right image of Fig.~\ref{fig:ActiveDescriptor} (a).

The camera model of panoramic annular lens is completely different from those of ordinary cameras where the pinhole model applies. Due to the wide FOV, the projection surface of the panoramic annular lens should be the curved surface (e.g. sphere), instead of the plane surface. Fortunately, the panoramic annular lens could be taken as the single-view wide-angle imaging system, and the related camera model has been studied by the research community~\cite{CalibrationPAMI, CalibrationECCV2004, CalibrationIROS, CalibrationICRA2007}. In this paper, OCamCalib~\cite{CalibrationIROS} toolbox is utilized to calibrate the panoramic annular lens and to obtain the intrinsic parameters. 

The unwrapping of the annular image is implemented complying with the following mapping, where the point $(i,j)$ of the rectangular image corresponds to the point $(x,y)$ of the annular image.
\begin{equation}
x = y_c + \rho\sin{\theta}~~~~~~~~~
y = x_c + \rho\cos{\theta}
\end{equation}
\begin{equation}
\rho = R_{min} + \frac{R_{max}-R_{min}}{height}i~~~~~~~~~
\theta = \frac{2\pi j}{width}
\end{equation}

The center of the annular image $(x_c, y_c)$ is estimated by the calibration toolbox. Subsequently, the circular borders of the annular image are manually determined [see the double circles in the left image of Fig.~\ref{fig:ActiveDescriptor} (a)], and $R_{max}$ and $R_{min}$ are thus obtained. The FOV ratio of the panoramic annular lens is utilized to determine the aspect ratio of the unwrapped rectangular image. The horizontal FOV of the panoramic annular lens is $360^\circ$, meanwhile the vertical FOV is determined by the projection model and the circular boarder of the annular image. In this paper, the panoramic annular lens features the vertical FOV of $75^\circ$, and the aspect ratio of unwrapped image is set to 4.8:1. 

\subsection{Active deep image descriptor}
In this paper, we leverage NetVLAD~\cite{netVLAD} to describe the panoramic images effectively. The NetVLAD model is based on the backbone network, which is usually the pre-trained network for the classification task on the large-scale image datasets (e.g. ImageNet~\cite{ImageNet} or Places~\cite{places}). As shown in Fig.~\ref{fig:ActiveDescriptor} (b), ResNet-18~\cite{resnet} without fully connected layers  is utilized as the base network of NetVLAD module. 

As an analogue of the pooling layer of the CNN, NetVLAD pools the discriminative descriptor with place fingerprint from the preceding feature map. 
The pooling ability of NetVLAD is obtained from the training phase, when the images with diverse appearances (e.g. with different illumination, viewpoint and dynamic objects) but captured at the same place are leveraged as training data. 
Specifically, the triplet loss function~\cite{netVLAD} impels the descriptor of the query image to close to that of positive database images rather than that of negative images, which robustifies the adaptability of the descriptor under variations.  
Similar with the training procedures in~\cite{netVLAD}, the dataset Pittsburgh-30k~\cite{netVLAD} with images of ordinary FOVs is utilized to in the training phase.


Having trained, the network is leveraged to extract  active deep image descriptors from panoramic images. 
As shown in Fig.~\ref{fig:ActiveDescriptor} (b), each unwrapped panoramic image is split into four parts along the horizontal direction. Subsequently, the four sub-images constituting a batch are fed into the proposed deep network to obtain four NetVLAD vectors. 
The active deep descriptor of the panoramic image is derived by adding (rather than concatenating) the four NetVLAD vectors, which is reasonable according to the principles of NetVLAD, meanwhile the final descriptor maintains compacted size.

\subsection{Sequential localization decision}
\begin{figure}[thpb]
	\centering
	\includegraphics[width=0.9\columnwidth]{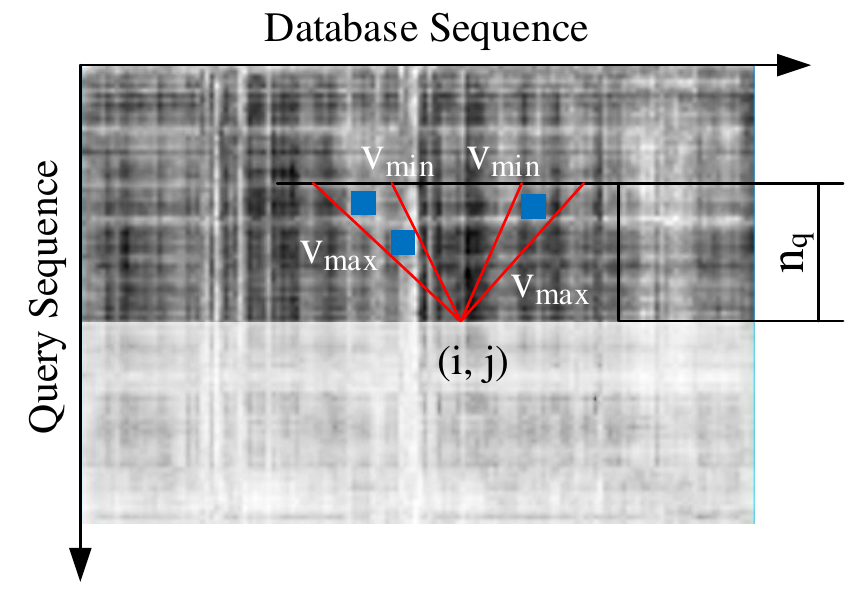}
	\caption{Using cone-based sequential searching to score the matching correspondence of images.}
	\label{fig:3}
\end{figure}

The extracted deep descriptors are leveraged to measure the similarity between images, thus to characterize the correspondence of query images and database images. 
Herein, we define $ D$ as the distance matrix, where the element $ D_{i,j} $ is the cosine distance between the \textit{i}-th query image and the \textit{j}-th database image. 
Inspired by the offline cone-based searching proposed in~\cite{OpenSeqSLAM2.0}, the online cone-based searching is executed upon the distance matrix $ D $. As shown in Fig.~\ref{fig:3}, each query-database pair $(i,j)$ within the distance matrix is associated with two symmetrical cone regions which is limited by sequential length $ n_{q} $, maximal velocity $ v_{max} $ and minimal velocity $ v_{min} $.  The online searching algorithm only makes use of the ``past'' query images instead of the ``future'' query images.

Within the regions, the number of best-matching pairs $ n_{match} $ is counted. The query descriptor and its nearest neighbor among the database descriptors compose the best-matching pair. The score of the query-database pair $(i,j)$ is defined as $ s_{i,j} = n_{match} / n_{q} $. Naturally, all of the matching scores form into a score matrix $ S $, where each query image corresponds to the best database image with the highest score. In order to get the final place recognition results, the matching score of the best query-database pair is evaluated through window uniqueness thresholding~\cite{OpenSeqSLAM2.0} to rule out the mismatched pairs. 

\section{Experiments}
In this section, both public and self-collected datasets are used to evaluate the proposed PAL system. Firstly, the performance of active deep descriptor was validated on MOLP dataset~\cite{MOLP}. Secondly, the visual localization performance was tested in the field. 

\subsection{Comparison between passive and active deep descriptors}
In the public panoramic dataset MOLP, four binocular cameras mounted on the vehicle were utilized to capture images of four different directions. 
In this paper, the summer night images of city roads captured in the reverse traversing direction are used to evaluate the performance of deep descriptors, including passive and active descriptors.

If the index difference between the place recognition result and the ground truth is not larger than the tolerance (set to $ 5 $ in this paper), the result is defined as a $ TP $ (true positive) result. Otherwise, the positive result is defined as a $ FP $ (false positive) result. Moreover, if the ground truth is not empty but the query does not match with any database image, the negative result is defined as a $ FN $ (false negative) result. The performance of localization is evaluated and analyzed in terms of $ F_1 $ score.
\begin{equation}
	 F_1 = 2 \times (P \times R)/(P + R),
\end{equation}
where  $ R = TP/(TP+FN) $ and $ P = TP/(TP+FP) $.

According to our previous research~\cite{VisualLocalizer}, the deep descriptors derived from GoogLeNet~\cite{GoogLeNet} pretrained on ImageNet~\cite{ImageNet} is the optimal choice on the task of visual localization among the prevailing networks. Therefore, the five optimal feature maps of convolutional or pooling layers in GoogLeNet (listed in Table~\ref{table_example}) are selected as one of the baseline of passive descriptors. Moreover, the conv3 and fc6 layer of AlexNet~\cite{AlexNet} pretrained on ImageNet are also used as another baseline of passive descriptors. Similar to the active descriptor extraction, the split images are fed into the network to obtain the feature maps of the designated layer, and those feature maps derived from sub-images are flatten and concatenated together to form into the descriptor. The reverse traversing direction of database and query sets is taken as a priori knowledge to concatenating the partial descriptors in the consistent order of sub-images.

Among active deep descriptors, the NetVLAD descriptors based on different backbone networks (i.e. AlexNet, VGG-16~\cite{vgg16} and ResNet-18) are compared to choose the optimal structure. The fully connected layers of the backbone networks are removed and the last convolutional layer is connected with the successive NetVLAD module. In this paper, all of the NetVLAD networks are trained on the public dataset Pittsburgh-30k with the default training parameters~\cite{netVLAD}. In order to compare various deep descriptors fairly, the input images fed into different networks are universally set to $ 224\times224 $. Meanwhile, the split number of panoramic images are also compared in the experiment, where four-part split, two-part split and one-part split are tried. The brute force searching is utilized to find the nearest neighbor of the query descriptor as the best-matching results, which are evaluated with $F_1$ score. 
\begin{table}[h]
\caption{The localization performance ($F_1$ score) of different split number on MOLP dataset (In.=Inception)}
\label{table_example}
\begin{center}
\begin{tabular}{|c|c|c|c|c|}
\hline
 \multicolumn{2}{| c |}{$F_1$} & One-part & Two-part & Four-part \\
\hline
\multirow{2}{*}{AlexNet}  & conv3&0.06&0.43&0.38\\
 & fc6&0.03&0.18&0.11\\
\hline
\multirow{5}{*}{GoogLeNet}  & In.3a/3$\times$3&0.04&0.29&0.26\\
 & In.3a/3$\times$3\_reduce&0.07&0.27&0.12\\
  & In.3b/3$\times$3\_reduce&0.06&0.51&0.37\\
   & In.3a/pool\_proj&0.07&0.34&0.25\\
      & In.5b/1$\times$1&0.09&0.32&0.06\\
\hline
 \multicolumn{2}{| c |}{NetVLAD with AlexNet}&0.19& 0.29&0.28\\
\hline
 \multicolumn{2}{| c |}{NetVLAD with VGG-16}&0.33& 0.51&0.50\\
\hline
 \multicolumn{2}{| c |}{\textbf{NetVLAD with ResNet-18}}&\textbf{0.41}& \textbf{0.51}&\textbf{0.54}\\
\hline
\end{tabular}
\end{center}
\end{table}

The experimental results are shown in Table~\ref{table_example}. The proposed active descriptors composed of NetVLAD network with ResNet-18 achieves the best performance among the listed descriptors. The feature maps of AlexNet and GoogLeNet behave better on the condition of two-part split, and are influenced substantially by the way of split. Comparatively, the descriptors derived from NetVLAD networks perform more stable among different split ways. Moreover, the superiority of the NetVLAD descriptors also lies in the combination way of sub-image descriptors. Due to the principle of netVLAD, the direct superposition of four descriptors does not require to know the traversing direction of datasets.  




\subsection{Performance on the real-world scenarios}
\begin{figure}[thpb]
	\centering
	\includegraphics[width=\columnwidth]{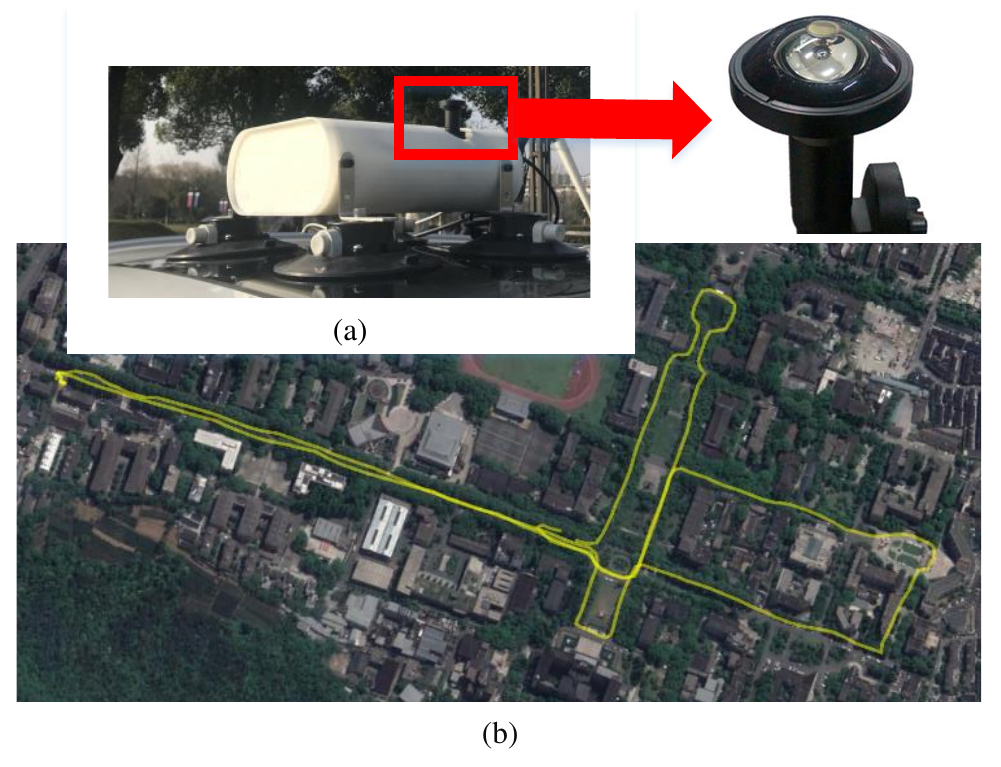}
	\caption{(a) The panoramic annular lens and the peripheral device are mounted on the top of the car. (b) The test route (denoted by yellow lines) covers around three kilometers.}
	\label{fig:hardware}
\end{figure}

The performance of PAL is validated on the Yuquan dataset, which was captured at the Yuquan Campus of Zhejiang University, China. The panoramic images were captured by the car-mounted panoramic annular lens on a three-kilometer route, as shown in Fig.~\ref{fig:hardware}. The database sequence were captured in the sunny afternoon, meanwhile the query sequences were captured both in the afternoon (subset-1) and at dusk (subset-2). Meanwhile, GNSS data were also collected.

\begin{figure}[thpb]
	\centering
	\subfigure[]{
		\includegraphics[width=\columnwidth]{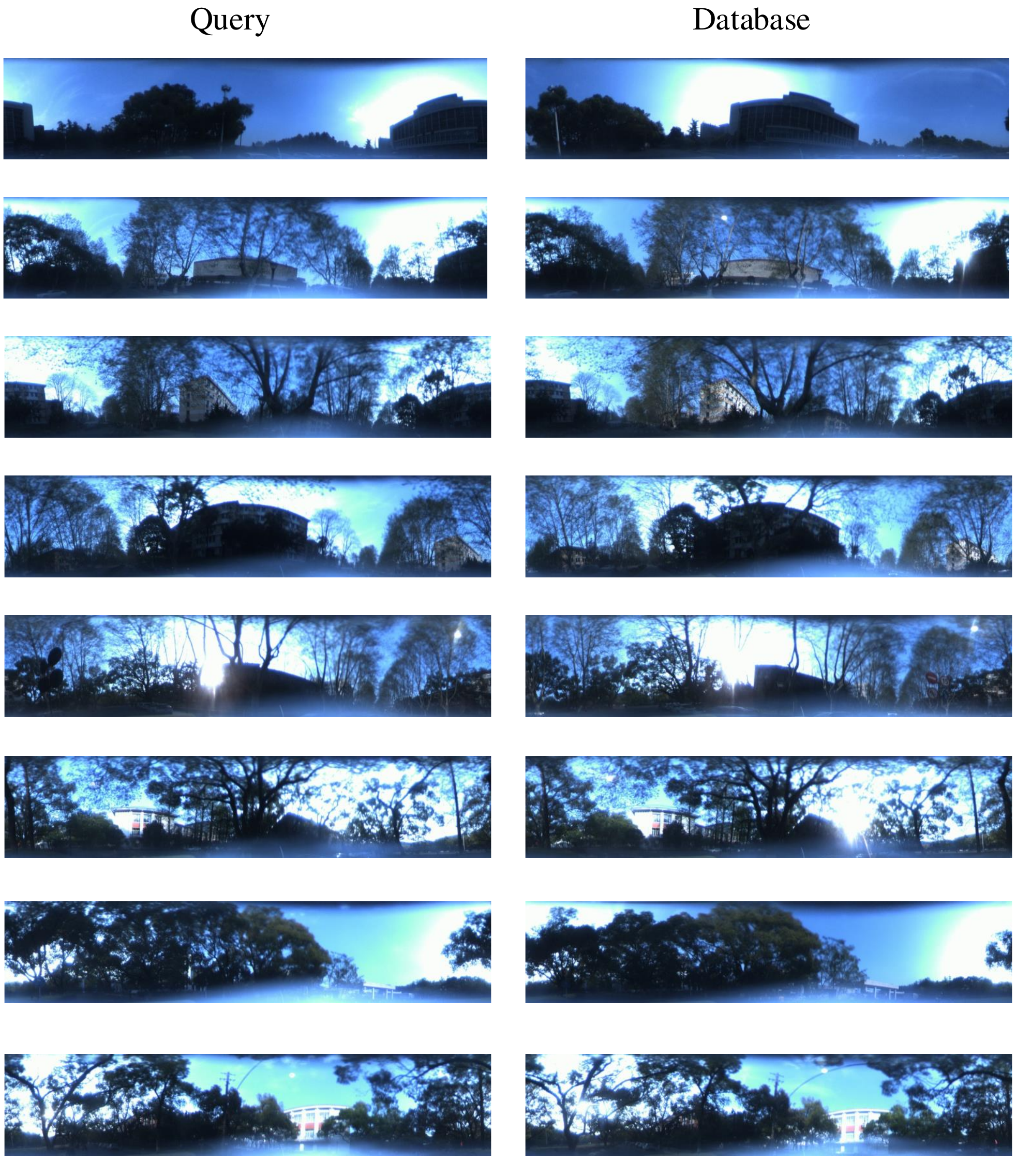}
	}
\subfigure[]{
	\includegraphics[width=\columnwidth]{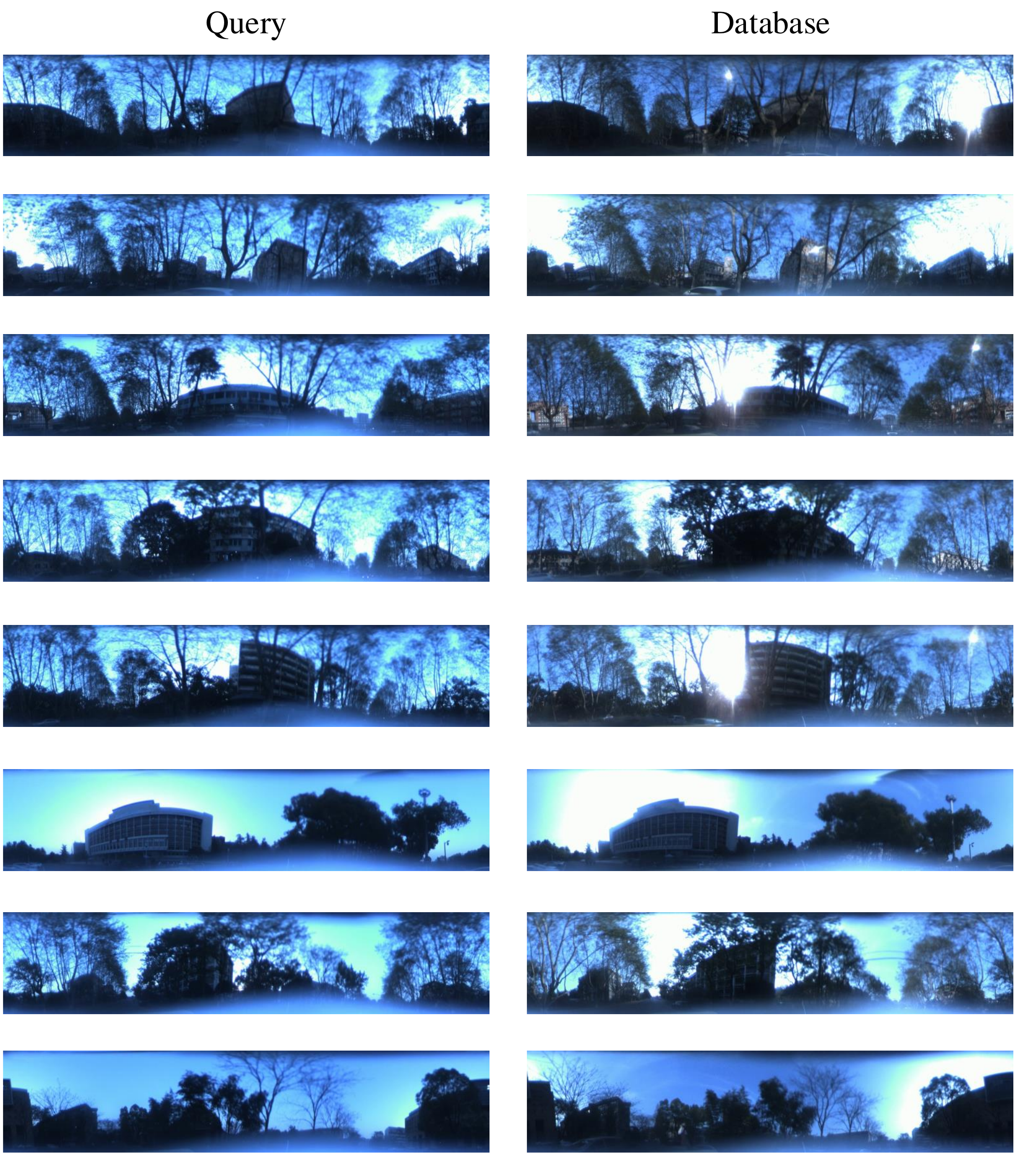}
	
}
	
	\caption{The visual localization results of (a) the afternoon test set and (b) the dusk test set.}
	\label{fig:afternoon}
\end{figure}

It is worthwhile to note that the database sequence is not completely overlapped with the two query sequences. Those unseen query images matched with the database image are denoted as false results. The ratio of false results out of all positive results is $ FR $ (false rate). The $ PR $ (positive rate) of localization is defined as the ratio of matching results, the distance between which is within the tolerance ($ 50 m $). In the experiment, the sequential matching parameters are set as $v_{min}=0.4$, $v_{max}=2.5$, $n_q=10$. As shown in Table~\ref{table_2}, the proposed PAL surpasses OpenSeqSLAM2.0~\cite{OpenSeqSLAM2.0} on the real-world dataset. Some localization results of PAL system are shown in Fig.~\ref{fig:afternoon}. It is concluded that the proposed approach is eligible to overcome the appearance variations (especially illumination variation). Moreover, the panoramic camera mounted on the roof of the car naturally reduces the impact of dynamic objects (like pedestrians and cars) on the performance of visual localization.

\begin{table}[h]
	\caption{The localization performance on Yuquan dataset}
	\label{table_2}
	\begin{center}
		\begin{tabular}{|c|c|c|c|c|}
			\hline
			\multirow{2}{*}{Algorithms }&  \multicolumn{2}{c|}{Subset-1}&\multicolumn{2}{c|}{Subset-2}\\
			\cline{2-5}
			&  $ FR$  & $ PR $ &$ FR $ & $ PR $  \\
			\hline			
			OpenSeqSLAM2.0  &17.15\%&21.67\%& 23.81\%&41.31\%\\
			\hline
			PAL  & \textbf{11.13\%}&\textbf{32.89\%}&\textbf{20.92\%}&\textbf{45.24\%}\\
			\hline
		\end{tabular}
	\end{center}
\end{table}


%

For computational complexity, the network inference of the deep feature extraction phase consumes around $ 40 ms $ on NVIDIA GPU (GTX-1060). Thereby, the active deep descriptor features not only robust description capability but also real-time performance. As for the phase of sequential matching, each image consumes around $13 ms$ to get the decision of place recognition. In a brief, the proposed PAL system could perform in real time in the practical scenarios.  
\section{CONCLUSIONS}

In order to tackling the variation issues of visual localization, PAL (panoramic annular localizer) is proposed in this paper. We incorporate panoramic annular lens and the active deep descriptor into the visual localization system. Firstly, the unwrapping of annular images is executed to prepare for the phase of descriptor extraction, where the split panoramic images are fed into the NetVLAD network based on ResNet-18. The descriptors obtained from sub-images are added together regardless of concatenation order, and the cone-based matching is leveraged to robustify the single-frame retrieval results. The experiment on MOLP dataset illustrate that the proposed active descriptor outperforms the off-the-shelf deep descriptors. In the field test, the performance of the proposed system in the practical scenarios is validated. 

The code and dataset related to the proposed PAL system are publicly available at https://github.com/chengricky/PAL. In the future, the scene understanding and pose estimation are planned to be researched based on this work.




%

\section*{ACKNOWLEDGMENT}
This work was supported by the State Key Laboratory of Modern Optical Instrumentation.

\bibliographystyle{IEEEtran.bst}
\bibliography{IEEEabrv.bib,mybibfile.bib}

\end{document}